# Extended Probabilistic Rand Index and the Adjustable Moving Window-based Pixel-pair Sampling Method

Hisashi Shimodaira

**Abstract**

*The probabilistic Rand (PR) index has the following three problems: It lacks variations in its value over images; the normalized probabilistic Rand (NPR) index to address this is theoretically unclear, and the sampling method of pixel-pairs was not proposed concretely. In this paper, we propose methods for solving these problems. First, we propose extended probabilistic Rand (EPR) index that considers not only similarity but also dissimilarity between segmentations. The EPR index provides twice as wide effective range as the PR index does. Second, we propose an adjustable moving window-based pixel-pair sampling (AWPS) method in which each pixel-pair is sampled adjustably by considering granularities of ground truth segmentations. Results of experiments show that the proposed methods work effectively and efficiently.*

## 1. Introduction

This paper deals with a criterion for quantitatively evaluating segmentations of images. Image segmentation is a process in which all the pixels of an image are classified into a number of regions. It is desired that the resultant regions correspond to constituent objects or their parts in the image. Moreover, it is implicitly expected that the resultant segmentation matches the interpretation of the image by human perception. Since perception of an image differs for each human, image segmentation and also quantitatively evaluating resultant segmentations is difficult.

Up to now, various criteria for quantitatively evaluating segmentations of images have been developed [1, 2, 3]. In this paper, we deal with an extension of the probabilistic rand (PR) index [2, 3] that is an application of the Rand index [4] enabling comparison of a test segmentation with the multiple ground-truth (GT) segmentations. The PR index is one of indices which have been widely used to compare image segmentation algorithms [5].

The Rand [4] index is a criterion for evaluating a measure of similarity between two clusterings of the same set of data using pairs of points. It is defined by considering how each pair of data points is classified in each cluster and counting the number of pairs representing their similarity. The Rand index takes values in [0, 1] according to the similarity between the two clusterings.

The PR index [2, 3] is a criterion for evaluating a measure of similarity between a test segmentation and the multiple GT segmentations which are generally produced by humans. It is defined by considering how each pixel-pair is classified in each segment (region) and by counting the number of pairs representing their similarity. Since computing it exhaustively over all pixel-pairs is computationally expensive, a limited number of pixel-pairs are sampled and used. The PR index takes values in [0, 1] according to the similarity between the test segmentation and the GR segmentations.

The PR index has the following three problems.

(1) It suffers from lack of variation in its value over images [2, 3], which is likely due to its smaller effective range.

(2) To remedy this, the authors of [2, 3] presented the normalized probabilistic Rand (NPR) index. It is defined by a simple linear transformation of the PR index using the expected value of the PR index. However, its theoretical foundation is unclear because the expected value of the PR index is formulated based on a theoretically unclear assumption. Also, the transformation is not necessary to compare the values of the index across images or algorithms, and computing the expected value of the PR index is computationally expensive.

(3) For the sampling method of pixel-pairs, although the authors of [3] described in Section 4, "in practice, we uniformly sample $5 \times 10^6$ pixel-pairs for an image size of $321 \times 481$ pixels instead of computing it exhaustively over all pixel-pairs," the sampling method was not explicitly described. We insist that the method of sampling pixel-pairs greatly affects the magnitude of the PR index as shown in Section 5, and the number of pixel-pair samples directly affects the computational cost. These problems are of crucial importance for the practical use of the PR index. To our best knowledge, these problems have not been described in any papers.

To solve these problems, we propose the following methods.

(1) The extended Rand (ER) index considering not only similarity but also dissimilarity between two clusterings. It takes values in [-1, 1] according to the similarity and dissimilarity between the two clusterings.

(2) The extended probabilistic Rand (EPR) index considering not only similarity but also dissimilarity between a test segmentation and the GT segmentations. In the same way as the PR index, it is formulated based on how each pixel-pair is classified in each region. It is also computed using a limited number of sampled pixel-pairs. It takes values in [-1, 1] according to the similarity and dissimilarity between a test segmentation and the GT segmentations. Note that the ER and EPR indices are formulated based on firm theoretical foundation, and the ranges of values that they can take are two times greater than those of the Rand and PR indices. This leads to magnifying variations of the values of the EPR index over segmentations.

(3) An adjustable moving window-based pixel-pair sampling (AWPS) method as follows: each pixel-pair is formed by a pixel at a corner of the window and a pixel located at a grid point set in the window; the side lengths of the window and the grid spacings are set to be directly proportional to the means of the side lengths of the circumscribed rectangles of all the regions in the GT segmentations for a test segmentation; and the total number of pixel-pairs is controlled by two parameters.

We experimentally show the effectiveness of the EPR index and AWPS method.

The remainder of this paper is organized as follows. Section 2 outlines the Rand index and proposes the ER index. Section 3 outlines the PR index and proposes the EPR index. Section 4 proposes the AWPS method and its algorithm. Section 5 presents and discusses results of the experiments. Finally, Section 6 concludes the paper.

## 2. Extended Rand index

### 2.1. Rand index

The Rand index uses pairs of data points and measures the similarity between clusterings based on how each pair of data points is assigned to clusters in each clustering. We consider the following two cases to represent a similarity between the clusterings: the two points of an individual point-pair are together assigned to a cluster in each of the two clusterings, and they are separately assigned to different clusters in both clusterings. From this, we count the number of assignments of point-pairs representing a similarity between the two clusterings of the same data, *Y* and *Y'*, and define the Rand index, *R(Y, Y')*, as the ratio of the number of similar assignments of point-pairs to the total number of point-pairs.

| Type | Point-pair | Total number |
|---|---|---|
| Together in both | *ab, de* | 2 |
| Separated in both | *ad, ae, af, bd, be, bf, cf* | 7 |
| Mixed | *ac, bc, cd, ce, df, ef* | 6 |

Table 1: Classification of point-pairs in clusters *Y* and *Y'*.

Consider the following example that calculates *R(Y, Y')* between two clusterings of six points. Let *Y* = {(*a, b, c*), (*d, e, f*)} and *Y'* = {(*a, b*), (*c, d, e*), (*f*)}. Table 1 shows the point-pairs. The total of nine similar assignments out of possible 15 point-pairs gives *R(Y, Y')* = 0.6.

More precisely, given $N_R$ points, $X_1, X_2, \cdots, X_{NR}$, and two clusterings $Y = \{Y_1, \cdots, Y_{K1}\}$ and $Y' = \{Y'_1, \cdots, Y'_{K2}\}$, we define the Rand index as

$$R(Y, Y') = \sum_{i<j}^{N} \gamma_{ij} / \binom{N_R}{2}, \quad (1)$$

where

$$\gamma_{ij} = \begin{cases} 1 & \text{if there exist } k \text{ and } k' \text{ such that both } X_i \text{ and } X_j \\ & \text{are in both } Y_k \text{ and } Y'_{k'}, \\ 1 & \text{if there exist } k \text{ and } k' \text{ such that } X_i \text{ is in both } Y_k \\ & \text{and } Y'_{k'}, \text{ whereas } X_j \text{ is in neither } Y_k \text{ nor } Y'_{k'}, \\ 0 & \text{otherwise.} \end{cases}$$

The value of the Rand index takes values in [0, 1], where 0 means the two clusterings have no similarities, and 1 means the clusterings are identical.

### 2.2. Extended Rand index

Here, we extend the Rand index by considering the following case not previously explicitly dealt with in it: the two points of a point-pair are together assigned to a cluster in one of the two clusterings, but they are separately assigned to different clusters in the other clustering. We consider this to represent dissimilarity between the clusterings. Then, we assign a value of 1 to a similar assignment of point-pairs and a value of -1 to a dissimilar assignment of point-pairs. From this, we define the ER index as the ratio of the total sum of the values of point-pairs to the total number of point-pairs.

For the example in Section 2.1, since the sum of the values of the similar assignments is 9 and that of the dissimilar assignments -6, the total is 3 and thus *ER* = 0.2.

Using the same notation as in Section 2.1, the ER index is defined as

$$ER(Y, Y') = \sum_{i<j}^{N} \gamma_{ij} / \binom{N_R}{2}, \tag{2}$$

where

$$\gamma_{ij} = \begin{cases} 1 & \text{if there exist } k \text{ and } k' \text{ such that both } X_i \text{ and } X_j \\ & \text{are in both } Y_k \text{ and } Y'_{k'}, \\ 1 & \text{if there exist } k \text{ and } k' \text{ such that } X_i \text{ is in both } Y_k \\ & \text{and } Y'_{k'}, \text{ whereas } X_j \text{ is in neither } Y_k \text{ nor } Y'_{k'}, \\ -1 & \text{if there exist } k \text{ and } k' \text{ such that } X_i \text{ is in both } Y_k \\ & \text{and } Y'_{k'}, \text{ whereas } X_j \text{ is in } Y_k \text{ but not in } Y'_{k'}, \\ & \text{or } X_j \text{ is in } Y'_{k'} \text{ but not in } Y_k. \end{cases}$$

From the theoretical perspective, the ER index is related to the Rand index as follows:

$$ER = 2R - 1. \tag{3}$$

The ER index takes values in [-1, 1], where -1 means the two clusterings are completely dissimilar, and 1 means the clusterings are identical. Note that the ER index is formulated based on firm theoretical foundation, and its range of values is two times greater than that of the Rand index.

## 3. Extended probabilistic Rand index

### 3.1. Probabilistic Rand index

We briefly describe the PR index [2, 3]. Consider an image $X = \{x_1, x_2, \cdots, x_{Np}\}$ consisting of $N_P$ pixels. Let $S_T$ be the segmentation of the image to be compared with the GT segmentation set $S_G = \{S_1, S_2, \cdots, S_K\}$. We denote the label of a pixel $x_i$ by $l_{i, ST}$ in the segmentation $S_T$ and by $l_{i, Sk}$ in the GT segmentation $S_k$. It is assumed that each label $l_{i, ST}$ takes one of $L_T$ values for $S_T$, and correspondingly $l_{i, Sk}$ can take one of $L_k$ values for the GT segmentation $S_k$. In [2, 3], the PR index is defined by modeling label relationships for each pair of pixels $(x_i, x_j)$ as follows:

$$PR(S_T, S_G) = \frac{1}{T} \sum_{\substack{i, j \\ i \neq j}} \left[ c_{ij} \bar{p}_{ij} + (1 - c_{ij})(1 - \bar{p}_{ij}) \right], \tag{4}$$

where

$$c_{ij} = II\left( l_{i, S_T} = l_{j, S_T} \right), \tag{5}$$

$$\bar{p}_{ij} = \frac{1}{K} \sum_{k} II\left( l_{i, S_k} = l_{j, S_k} \right), \tag{6}$$

where $T$ is the total number of sampled pixel-pairs and $II$ is the identity function.

The PR index takes values in [0, 1], where 0 means the test segmentation and the GT segmentations have no similarities, and 1 means that the test segmentation and all the GR segmentations are identical.

### 3.2. Extended probabilistic Rand index

The EPR index is an extension of the PR index based on the same theoretical foundation as the ER index. Using the same notation as in Section 3.1, we define the EPR index as follows:

$$EPR(S_T, S_G) = \frac{1}{T} \sum_{\substack{i, j \\ i \neq j}} \left[ F\left( l_{i, S_T}, l_{j, S_T} \right) \bar{F}\left( l_{i, S_G}, l_{j, S_G} \right) \right], \tag{7}$$

where

$$\bar{F}\left( l_{i, S_G}, l_{j, S_G} \right) = \frac{1}{K} \sum_{k} F\left( l_{i, S_k}, l_{j, S_k} \right), \tag{8}$$

and we define the function $F$ for arbitrary positive integers $m$ and $n$ as

$$F(m, n) = \begin{cases} 1 & \text{if } m = n \\ -1 & \text{if } m \neq n \end{cases}. \tag{9}$$

Note that this index considers the dissimilar assignments of pixel-pairs as in the ER index.

From the theoretical perspective, the PR and EPR indices computed using the same pixel-pairs have the following relationship:

$$EPR = 2PR - 1. \tag{10}$$

The EPR index takes values in [-1, 1], where -1 means the test segmentation and the GT segmentations have no similarities, and 1 means the test segmentation and all the GT segmentations are identical. Note that the EPR index is formulated based on firm theoretical foundation, and its range of values is two times greater than that of the PR index.

The EPR index is decomposed into three terms, each of which is a summation of elements in Eq. (7) classified according to the combinations of the signs of the function values of the test segmentation (*TF*) and those of the GT segmentations (*GF*). That is,

$$EPR = RPP + RMM + RPM, \tag{11}$$

where *RPP* is the term for the case where both *TF* and *GF* have positive signs, *RMM* where both *TF* and *GF* have negative signs, and *RPM* where *TF* and *GF* have different signs from each other. Note that *RPP* and *RMM* represent the similarity between the test and GT segmentations, and the *RPM* represents the dissimilarity.

The computational complexities for the PR and EPR indices are as follows. Those for (6) and (8) are the same. The computational complexity per one sampling point for (4) is higher than that for (7) by the second term in the square brackets (two subtractions and one multiplication).

## 4. Method and algorithm for sampling pixel-pairs

### 4.1. Fundamental principle for sampling pixel-pairs

Consider a simple example shown in Figure 1, which represents a test segmentation (broken line) and the GT segmentation (solid line) each of which consists of a region of interest and the background region. Considering a pixel P inside both regions of interest, we have three types of pixel-pairs corresponding to the location of the counterpart pixel Q: P-$Q_1$, P-$Q_2$, and P-$Q_3$. These correspond to the three types of point-pairs described in Section 2.2, and the results of evaluation by these correspond to the three terms of Eq. (11). That is, P-$Q_1$ evaluates the overlapping area between the two regions of interest, and the result corresponds to RPP; P-$Q_2$ evaluates the overlapping area between the two background regions, and the result corresponds to RMM; and P-$Q_3$ evaluates the non-overlapping area between the two regions of interest, and the result corresponds to RPM. Thus, P-$Q_1$ and P-$Q_2$ evaluate the similarity between the two segmentations and the P-$Q_3$ evaluates the dissimilarity.

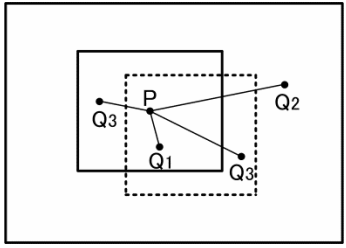

Figure 1: Schematic diagram showing notion of the three types of pixel-pairs (P-$Q_1$, P-$Q_2$, and P-$Q_3$) for comparing a test segmentation (broken line) with the GT segmentation (solid line).

The above consideration shows that essential properties of a pixel-pair to be noted are the span between the two pixels and the direction formed by them. From this example, we see the following: to evaluate the overlapping and non-overlapping areas between the two regions of interest accurately, the pixel-pairs should have appropriate spans and be in all directions, and we need to sample such pixel-pairs in an appropriate proportion. Moreover, since the benchmark of comparison is the GT segmentation, they should be sampled according to the sizes of regions in the GT segmentation.

This gives us the following fundamental principle to sample pixel-pairs: since actual GT segmentations consist of regions of various sizes, we need to sample pixel-pairs according to some measure representing the granularities of the GT segmentations. To compare the overlapping and non-overlapping areas of all regions, the sampling should be uniformly performed over all the test and GT segmentations. To evaluate various regions of various shapes, pixel-pairs with appropriate spans should be selected uniformly in all directions without duplicates. In the sampling procedure, randomness is not appropriate

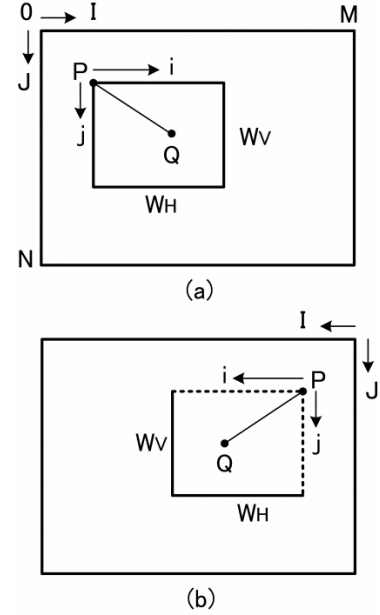

Figure 2: Window system for sampling pixel-pairs.

because it makes the evaluation results unreproducible.

## 4.2. Method for sampling pixel-pairs

Based on the aforementioned fundamental principle, we design the adjustable moving window-based pixel-pair sampling (AWPS) method as follows. As the measure representing the granularities of actual GT segmentations in the horizontal and vertical directions, we use the means of the widths and heights of the circumscribed rectangles of all regions in all GT segmentations for a test segmentation of interest. We denote these by $H_m$ and $V_m$, respectively. We set the window system as in Figure 2 and set the width $W_H$ and height $W_V$ of the window to be directly proportional to $H_m$ and $V_m$ as

$$\left.\begin{array}{l} W_H = \alpha\, H_m \\ W_V = \alpha\, V_m \end{array}\right\}, \qquad (12)$$

where $\alpha$ is a proportional constant. Then, we set a grid in the window whose horizontal and vertical spacings, $d_H$ and $d_V$, are set to be directly proportional to $H_m$ and $V_m$ as

$$\left.\begin{array}{l} d_H = \beta\, H_m \\ d_V = \beta\, V_m \end{array}\right\}, \qquad (13)$$

where $\beta$ ($\beta < \alpha$) is a proportional constant. The optimum values of $\alpha$ and $\beta$ are determined experimentally. Each pixel-pair is formed by pixels P and Q, where P is located at the upper left or upper right corner of the window, and Q is located at one of the grid points. This window system is moved from the left to the right and the right to the left to sample pixel-pairs.

## 4.3. Algorithm for sampling pixel-pairs

Here, we present the algorithm for implementing the above method. Let $M$ and $N$ be the horizontal and vertical sizes of an image, respectively. Figure 3 shows the pseudocode for when the window is moved from the left to the right as in Figure 2(a). Note that $J$ and $I$ are taken as even numbers. The code for when the window is moved from the right to the left in Figure 2(b) is implemented in a similar way; but in this case, $J$ and $I$ are taken as odd numbers, and grid points on the dotted lines are omitted from candidate points for Q because pixel-pairs in horizontal and vertical directions are sampled when the window is moved from the left to the right. In both cases, when $j = J$ and $i = I$, the pixel-pair is omitted. Further, non-overlapping portions of the window at the periphery of an image are omitted. In this way, a pixel-pair is selected and used to partially compute the values of $EPR$ in Eq. (7). Note that pixel-pairs are selected uniformly in all directions without duplicates. Pixel-pairs sampled are in a rectangle as in Figure 4, where P is a pixel and Q is the counterpart pixel of the pixel-pair because P and Q are interchangeable with each other.

The total number of pixel-pairs sampled by this method, $N_T$, is approximately represented as follows:

$$N_T \cong \frac{1}{4}\left\{\left(\frac{\alpha}{\beta}+1\right)^2 + \left(\frac{\alpha}{\beta}\right)^2\right\} MN. \qquad (14)$$

```
for (J = 0; J < N; J += 2)
    for (I = 0; I < M; I += 2)
        for (j =J; j < J + W_V; j += d_V)
            for (i =I; i < I + W_H; i += d_H)
                P ←(J, I)
                Q ←(j, i)
            End i
        End j
    End I
End J
```

Figure 3. Pseudocode of pixel-pair sampling algorithm for when the window is moved from the left to the right.

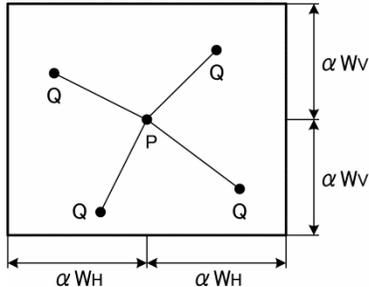

Figure 4: Area in which counterpart pixels Q of pixel-pairs for a pixel P are located.

In this equation, non-overlapping portions of the window at the periphery of an image are neglected. Irrespective of the granularities of GT segmentations, $N_T$ is a function of only the ratio $\alpha / \beta$ for images of the same size. Note that the AWPS method has the following desirable properties: the window size and sampling spacing are adjustable according to the granularities of GT segmentations, and $N_T$ does not depend on GT segmentations although it slightly changes due to the incorrectness at the periphery of an image. $N_T$'s for images with the same size are almost the same if the same values of $\alpha$ and $\beta$ are used.

## 5. Experiments and discussions

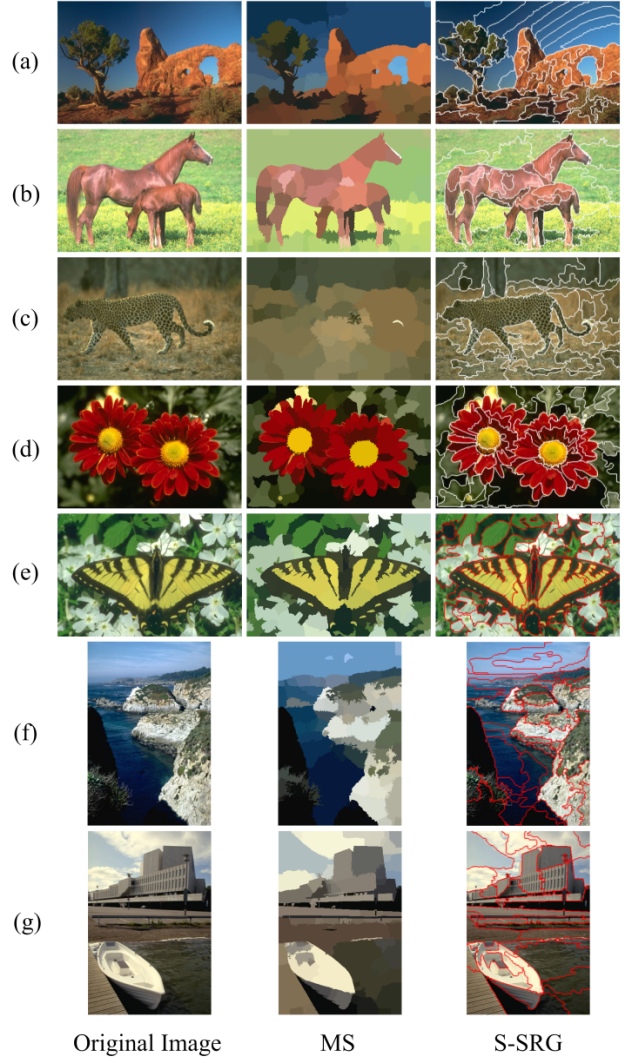

Figure 5: Test images and test segmentations produced by MS and S-SRG: (a) rock, (b) horses, (c) leopard, (d) flowers, (e) butterfly, (f) coast, and (g) boat.

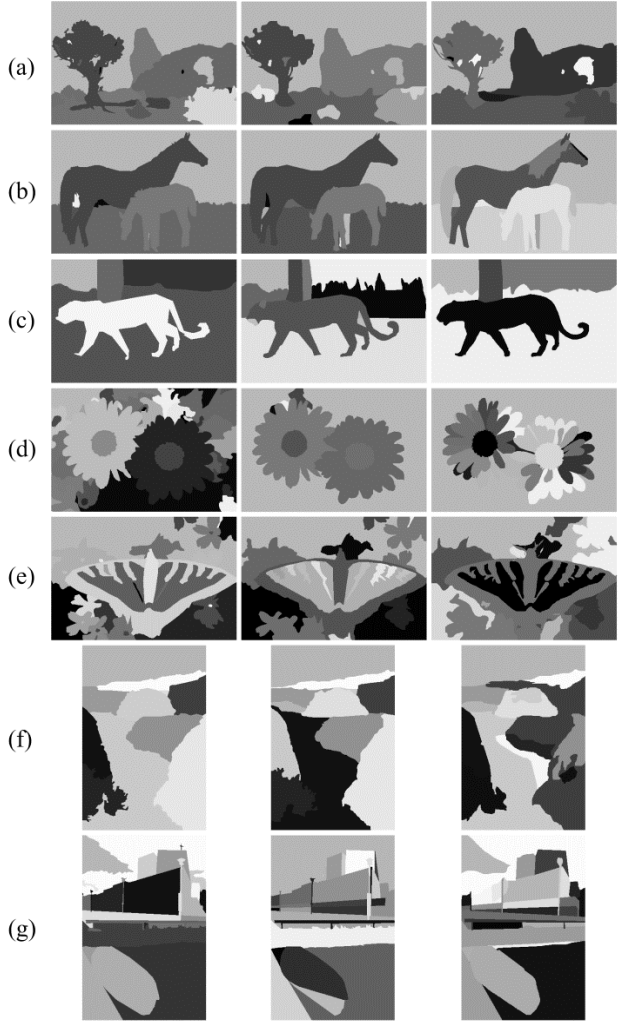

Figure 6: GT segmentations: (a) rock, (b) horses, (c) leopard, (d) flowers, (e) butterfly, (f) coast, and (g) boat.

| image | rock | horses | leopa. | flowers | butter. | coast | boat |
|---|---|---|---|---|---|---|---|
| $Hm$ | 99.43 | 140.9 | 230.3 | 82.33 | 100.1 | 152.2 | 93.43 |
| $Vm$ | 62.10 | 91.22 | 123.3 | 71.72 | 66.12 | 126.3 | 61.79 |

Table 2: Values of $Hm$ and $Vm$ in pixels of GT segmentations for test images.

We performed experiments to investigate the effectiveness of the proposed methods. In the experiments, for the test segmentations, we used segmentations produced by the mean shift (MS) [6] and the automatic image segmentation method using square elemental region-based seeded region growing and merging method (S-SRG) [7]; and for the GT segmentations, we used segmentations in the database of human segmented natural images [8]. We show the main results of the experiments. Figure 5 shows the seven test images and test segmentations produced by MS and S-SRG with the estimated best values of parameters: we used ($hs$, $hr$) = (16, 19) and a minimum region area of 80 for MS, and the parameter values described in [7] for S-SRG. The image size is 481 × 321 pixels or 321 × 481 pixels. Figure 6 shows the three GT segmentations used for each test image. Table 2 shows the values of $Hm$ and $Vm$ in pixels computed from the three GT segmentations.

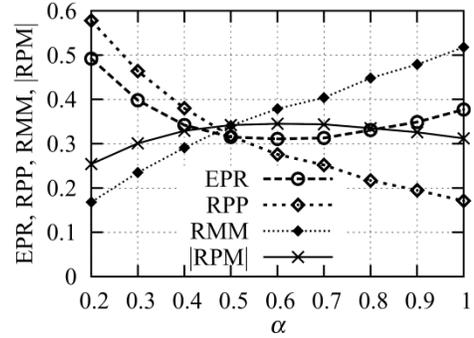

Figure 7: Variations in values of EPR, RPP, RMM, and RPM with values of α under β = 0.1α for test segmentation of horses image produced by MS.

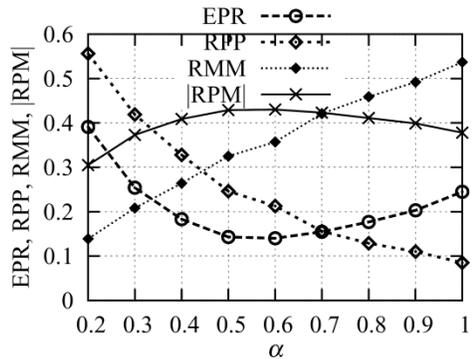

Figure 8: Variations in values of EPR, RPP, RMM, and RPM with values of α under β = 0.1α for test segmentation of flowers image produced by MS.

First, we performed experiments to determine the optimum value of α in Eq. (12). Under the condition of β = 0.1α, we investigated the variations in the values of *EPR*, and in those of *RPP*, *RMM*, and *RPM* in Eq. (11) with the values of α for the test segmentations of the horses and flowers images produced by MS. Figures 7 and 8 show the results. We can interpret the results intuitively by analogy with the example shown in Figure 1 as follows.

(1) *RPP* and *RMM* are the monotonic decreasing and increasing functions of α, respectively, and EPR becomes the minimum values at α = 0.6. We can see this as follows. When α is lower, many short-span pixel-pairs such as P-$Q_1$ are sampled, and this causes *RPP* to increase, which in turn leads *EPR* to increase. Alternatively, when α is higher, many long-span pixel-pairs such as P-$Q_2$ are sampled, this causes RMM to increase, which in turn leads EPR to increase. These results suggest that the optimum value of $\alpha$ exists.

| image | rock | hors. | leop. | flow. | butt. | coast | boat | mean |
|---|---|---|---|---|---|---|---|---|
| MS | 0.6 | 0.6 | 0.4 | 0.6 | 0.5 | 0.3 | 0.6 | 0.5143 |
| S-SRG | 0.7 | 0.8 | 0.5 | 0.6 | 0.5 | 0.4 | 0.6 | 0.5857 |

Table 3: Values of $\alpha_m$ for test segmentations produced by MS and S-SRG.

| $\beta$ | 0.0275 | 0.055 | 0.11 | 0.165 | 0.22 |
|---|---|---|---|---|---|
| $N_T$ | $2.10\times10^7$ | $6.50\times10^6$ | $1.94\times10^6$ | $7.79\times10^5$ | $3.93\times10^5$ |
| EPR | 0.3126 | 0.3122 | 0.3066 | 0.3006 | 0.2912 |

Table 4: Variations in values of $N_T$ and EPR with values of $\beta$ under $\alpha = 0.55$ for test segmentation of horses image produced by MS.

| $\beta$ | 0.0275 | 0.055 | 0.11 | 0.165 | 0.22 |
|---|---|---|---|---|---|
| $N_T$ | $3.03\times10^7$ | $6.21\times10^6$ | $1.69\times10^6$ | $8.23\times10^5$ | $4.15\times10^5$ |
| EPR | 0.1413 | 0.1417 | 0.1384 | 0.1265 | 0.1147 |

Table 5: Variations in values of $N_T$ and EPR with values of $\beta$ under $\alpha = 0.55$ for test segmentation of flowers image produced by MS.

(2) We note that absolute value of RPM (|RPM|) becomes the maximum at a value of $\alpha$ ($\alpha_m$), and EPR becomes the minimum at $\alpha_m$. For both images, the values of $\alpha_m$ are 0.6. We can see this as follows. By the use of $\alpha_m$, a relatively large number of pixel-pairs with spans appropriate for the occasion such as P-$Q_3$ are sampled, and the boundary areas of the GT segmentation are most closely evaluated by these pixel-pairs. Consequently, |RPM| reaches its highest value, and EPR reaches its lowest value.

The above consideration shows that by the use of $\alpha_m$, the difference between a test segmentation and the GT segmentation is most strictly evaluated, and consequently EPR becomes its lowest value. Since EPR is a criterion to measure the difference between a test segmentation and the GT segmentations, the use of $\alpha_m$ is desirable for the objective of EPR itself. If we can observe that such a characteristic is common in other test segmentations produced by MS and other algorithms, this gives foundation for using $\alpha_m$ as the optimum value of $\alpha$. This is preferable to avoid the arbitrariness of the EPR value from using an arbitrary sampling method.

Thereupon, we performed experiments to investigate the value of $\alpha_m$ under the condition of $\beta = 0.1\alpha$ for the test segmentations produced by MS and S-SRG. Table 3 shows the results. Although the values of $\alpha_m$ vary slightly, they are in a relatively small range centered at 0.5 or 0.6. These results provide an implicit rationale for the AWPS method. Thus, we consider $\alpha_m$ the optimum value of $\alpha$. Since the mean of the values of $\alpha_m$ for MS and S-SRG is 0.55, we used $\alpha = 0.55$ in experiments hereafter.

Next, we performed experiments to investigate the optimum value of $\beta$ in Eq. (13). Under the condition $\alpha = 0.55$, we investigated variations in the values of $N_T$ and EPR with values of $\beta$ for the test segmentations of the horses and flowers images produced by MS. Tables 4 and 5 show the results. Both results of EPR show the following: when the value of $\beta$ becomes larger than 0.055 (0.1$\alpha$), the values of EPR become significantly smaller gradually; whereas, when the value of $\beta$ becomes considerably smaller than 0.055, the value of EPR hardly varies. These results show that the optimum value of $\beta$ is 0.055 under the condition $\alpha = 0.55$ from the accuracy of the values of EPR. The mean of the values of $N_T$ for $\beta = 0.055$ is $6.36 \times 10^6$, and this value is almost equal to the value $5 \times 10^6$ recommended in [3]. It is approximately 0.06% of the total number ($1.125 \times 10^{10}$) of the possible pixel-pairs for an image size of 481 × 321 pixels. From these results, we determined $\alpha = 0.55$ and $\beta = 0.055$ as their optimum values and used them in experiments hereafter.

Second, we computed the values of PR and EPR using $\alpha = 0.55$ and $\beta = 0.055$ for the test segmentations produced by MS and S-SRG (see Figure 5). Tables 6 and 7 show the results. Table 8 shows the numbers of the regions for the test segmentations and the means of the numbers of the regions for the GT segmentations. For the values of EPR and PR, the relationship in Eq. (10) holds. The values of EPR present sufficient differences in values for comparing each other and appear to correspond to their visual evaluations for the test segmentations. The mean of the EPR values for S-SRG is slightly higher than that for MS. This also appears to correspond to their visual evaluations.

| image | rock | horses | leopa. | flowers | butter. | coast | boat | mean |
|---|---|---|---|---|---|---|---|---|
| MS | 0.6252 | 0.6561 | 0.5155 | 0.5708 | 0.6232 | 0.6959 | 0.7104 | 0.6282 |
| S-SRG | 0.5805 | 0.6470 | 0.5767 | 0.6399 | 0.6794 | 0.6413 | 0.7252 | 0.6414 |

Table 6: Values of PR using $\alpha = 0.55$ and $\beta = 0.055$ for test segmentations produced by MS and S-SRG.

| image | rock | horses | leopa. | flowers | butter. | coast | boat | mean |
|---|---|---|---|---|---|---|---|---|
| MS | 0.2504 | 0.3122 | 0.0311 | 0.1417 | 0.2463 | 0.3919 | 0.4209 | 0.2563 |
| S-SRG | 0.1610 | 0.2941 | 0.1535 | 0.2797 | 0.3588 | 0.2827 | 0.4504 | 0.2829 |

Table 7: Values of EPR using $\alpha = 0.55$ and $\beta = 0.055$ for test segmentations produced by MS and S-SRG.

| image | rock | horses | leopa. | flower | butter. | coast | boat |
|---|---|---|---|---|---|---|---|
| MS | 117 | 97 | 75 | 157 | 152 | 90 | 88 |
| S-SRG | 43 | 43 | 42 | 52 | 54 | 35 | 36 |
| GT | 22.0 | 13.0 | 7.7 | 34.7 | 29.7 | 13.0 | 35.7 |

Table 8: Numbers of regions for test segmentations produced by MS and S-SRG, and means of the numbers of regions for GT segmentations.

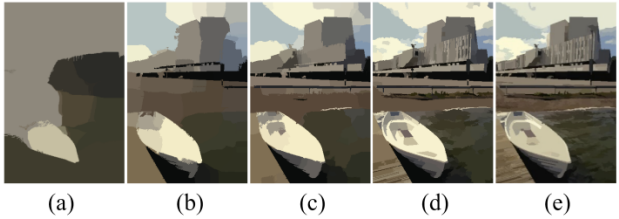

(a)　　　(b)　　　(c)　　　(d)　　　(e)

Figure 9: Test segmentations with different granularities produced by MS for boat image: (a) boat1, (b) boat2, (c) boat, (d) boat3, and (e) boat4.

| Image | boat1 | boat2 | boat | boat3 | boat4 |
|---|---|---|---|---|---|
| $N_r$ | 14 | 54 | 88 | 265 | 461 |
| EPR | 0.1562 | 0.3132 | 0.4209 | 0.2678 | 0.1598 |

Table 9: Numbers of regions ($N_r$) and values of EPR for test segmentations of boat image produced by MS.

Finally, we performed experiments to investigate the behavior of values of EPR for different segmentation granularities. For the boat image, we produced the under-segmented and over-segmented images shown in Figure 9 by MS and computed the values of EPR using $\alpha = 0.55$ and $\beta = 0.055$. Table 9 shows the results, where $N_r$ denotes the total number of the regions for a test segmentation. Note that compared to the best segmentation (boat), the values of EPR vary with the granularities of the segmentations and consequently present sufficient differences in values to judge what a good segmentation is.

The results of the experiments are summarized as follows:
- The AWPS method can sample an appropriate number of pixel-pairs efficiently using the optimum values of the parameters determined experimentally ($\alpha = 0.55$ and $\beta = 0.055$).
- The EPR index presents a wider range of values than the PR index and sufficient differences in values for comparing different segmentations.
- Its values vary with granularities of segmentations and provide sufficient differences in the values to allow judging what a good segmentation is.

The features of the EPR index are as follows:
- The EPR index is a linear transformation of the PR index like the NPR index. However, the former has clear theoretical foundation of the explicit consideration of dissimilarity, whereas the latter lacks clear theoretical foundation because it uses the expected value of the PR index based on a theoretically unclear assumption.
- The EPR index provides twice as wide effective range as the PR index does in evaluating segmentation results.
- The computational complexity per one sampling point for the EPR index is significantly lower than that for the PR index.

The features of the AWPS method are as follows:
- It samples each pixel-pair according to granularities of ground truth segmentations adjustably. To evaluate various regions of various shapes, pixel-pairs with appropriate spans are selected uniformly in all directions without duplicates.
- The parameter values can be determined so as to evaluate the difference between a segmentation and the GT segmentations strictly.
- The total numbers of sampled pixel-pairs for images with the same size are almost the same if the same parameter values are used. Since the optimal number of sampled pixel-pairs is 0.06% of the number of possible pixel-pairs, its computational cost is very low.
- If the user uses the same parameter values, the evaluation results are reproducible.

To our best knowledge, the PR index has been used only to compare image segmentation algorithms. Since the EPR index using the AWPS method can evaluate the difference between a segmentation and the GT segmentations strictly, we will use it as part of objective function for a segmentation algorithm, such as semantic segmentation [9, 10], to improve the performance.

## 6. Conclusions

In this paper, we have presented the EPR index and AWPS method. The results of the experiments have shown the following: the AWPS method is an effective and efficient sampling method, and the EPR index can present a wider range of values than the PR index and thus is a more effective criterion for evaluating quantitatively segmentations of images.

**Hisashi Shimodaira** received the BE, ME and DE degrees from Tokyo Metropolitan University in 1969, 1971, and 1982, respectively. He had been professor in the Information and Communications Department of Bunkyo University at Chigasaki City in Japan and retired from it in 2012. At present, he has no affiliations. His research interests include computer vision and artificial intelligence. Email: hshimodaira@c06.itscom.net